# Robust Kernel Sparse Subspace Clustering

Ivica Kopriva

Division of Electronics, Ruđer Bošković Institute, Zagreb, Croatia

*Abstract*—Kernel methods are applied to many problems in pattern recognition, including subspace clustering (SC). That way, nonlinear problems in the input data space become linear in mapped high-dimensional feature space. Thereby, computationally tractable nonlinear algorithms are enabled through implicit mapping by the virtue of kernel trick. However, kernelization of linear algorithms is possible only if square of the Froebenious norm of the error term is used in related optimization problem. That, however, implies normal distribution of the error. That is not appropriate for non-Gaussian errors such as gross sparse corruptions that are modeled by $\ell_1$-norm. Herein, to the best of our knowledge, we propose for the first time robust kernel sparse SC (RKSSC) algorithm for data with gross sparse corruptions. The concept, in principle, can be applied to other SC algorithms to achieve robustness to the presence of such type of corruption. We validated proposed approach on two well-known datasets with linear robust SSC algorithm as a baseline model. According to Wilcoxon test, clustering performance obtained by the RKSSC algorithm is statistically significantly better than corresponding performance obtained by the robust SSC algorithm. MATLAB code of proposed RKSSC algorithm is posted on https://github.com/ikopriva/RKSSC.

## I. INTRODUCTION

High dimensional data analysis is generic problem relevant for many application domains such as computer vision, machine learning, bioinformatics, pattern recognition, signal processing, medical image analysis, to name a few [1]. Thereby, clustering or partitioning high dimensional data into disjoint homogeneous groups is one of the fundamental problems in data analysis [2]. It aims to infer structure from data based on similarity between data points. However, sample spaces often have arbitrary (complex) shape and distance-based clustering algorithms fail to cluster data in the original ambient domain. Furthermore, high-dimensionality of the ambient domain deteriorates performance even further, and that is related to the well-known phenomenon of the *course of dimensionality*. Consequently, identification of low-dimensional structure of data in high-dimensional ambient space is one of the fundamental problems in fields of engineering and mathematics [1]. Nevertheless, in many applications, data are well represented by a union of multiple linear subspaces giving rise to linear subspace clustering (SC) [3]-[7]. Unfortunately, in real world data do not necessarily come from linear subspaces. One way to address such problem is formulation of SC algorithms, instead in the original input space $\mathcal{X}$, in reproducible kernel Hilbert space (RKHS) $\mathcal{H}$, a.k.a the feature space. It is induced by mapping $\phi:\mathcal{X} \to \mathcal{H}$, that is justified by the Cover's theorem [8]. It states that number of separating hyperplanes is proportional to dimension of data generating space. Thus, it is important for dimension of $\mathcal{H}$ to be large. In that case, data will be linearly separable.

However, working in high (infinite) dimensional feature space is computationally intractable. Standard solution for this problem is employment of the *kernel trick*, [9]-[11]. There are, however, two unresolved issues with kernel-based methods: (*i*) linear SC algorithms that do not use the Froebenious norm of the error term, such as robust version of the sparse SC (SSC) [5], cannot be kernelized [12]; (*ii*) after many years of research it is still unclear how to choose the kernel function such that empirical data fit kernel-induced RKHS. Herein, we focus on the first issue.

Kernelization of linear algorithms, not only SC algorithms, is possible if model error, a.k.a. data fidelity term, in related optimization problem is expressed in terms of the square of the $\ell_2$-norm when dealing with vectors, or in terms of the square of the Froebenious norm when dealing with matrices. That, however, implies normal distribution of the error term and that is incorrect for gross sparse corruptions and outliers that are modeled by the $\ell_1$-norm of the error term. Kernelization that takes into account outliers is accomplished in [10] and [11] for $\ell_{2,1}$-norm of the error term. As pointed out in [12], it is still an open problem how to kernelize linear algorithms based on the $\ell_1$-norm of the error term that models gross sparse corruptions.

Instead of applying linear algorithms in induced high (possibly infinite) dimensional space, it is possible to do that in the empirical feature space [13]. It is induced by approximate explicit feature map that it includes only limited number of terms from $\phi$ [14], or by empirical kernel map (EKM) [15]. In particular, the kernel principal component analysis (KPCA) transform [16] is actually a special type of EKM [17][13]. That EKM was called nonlinear projection trick (NPT) in [18]. If data in feature space are centered, NPT enables calculation of coordinates of mapped data with respect to orthonormal basis of a subspace of the empirical feature space. Then, applying the linear method to the coordinates of mapped data is equivalent to applying the kernel method to the original data [18]. In [18] rigorous proofs of equivalence were presented for KPCA, kernel support vector machines, kernel Fischer discriminant analysis and for kernel-based version of the robust principal component analysis, i.e. PCA-L1 [19]. As opposed to other three algorithms, it is not possible to obtain kernel PCA-L1 by using the *kernel trick*. Herein, we present a rigorous proof of equivalence between linear robust, $\ell_1$-norm based, SSC algorithm and its NPT-based kernel version.

The rest of the paper is organized as follows. In Section II we revisit a background and related work. We present NPT-based kernel SSC algorithm in Section III. Experimental results are presented in Section IV, while conclusions are presented in Section V.

## II. BACKGROUND AND RELATAED WORK

### A. Empirical kernel map

Let $\mathbf{X} \triangleq [\mathbf{x}_1,...,\mathbf{x}_N] \in \mathbb{R}^{D \times N}$ represent in-sample (training) dataset comprised of $N$ samples in $D$-dimensional input space. Let $\Phi(\mathbf{X}) \triangleq [\phi(\mathbf{x}_1),...,\phi(\mathbf{x}_N)] \in \mathbb{R}^{F \times N}$ stand for in-sample data in the $F$-dimensional feature space.

**Definition 1** [15]. For a given in-sample dataset $\mathbf{X} \triangleq [\mathbf{x}_1,...,\mathbf{x}_N]$, we call:

$$\phi_N : \mathbb{R}^D \to \mathbb{R}^N, \mathbf{x} \mapsto k(.,\mathbf{x})\big|_{(\mathbf{x}_1,...,\mathbf{x}_N)} \\ = [k(\mathbf{x}_1,\mathbf{x}),...,k(\mathbf{x}_N,\mathbf{x})] \quad (1)$$

the empirical kernel map with regard to $[\mathbf{x}_1,...,\mathbf{x}_N]$.

The EKM induces RKHS $\mathcal{H}_N \subset \mathcal{H}$ that is endowed with the inner product $\langle \cdot, \cdot \rangle$ such that $k(\mathbf{x},\mathbf{y}) = \langle \phi_N(\mathbf{x}), \phi_N(\mathbf{y}) \rangle \in \mathbb{R}$, i.e. the *kernel trick*.

### B. Nonlinear projection trick

Herein, we briefly review the NPT concept [18]. Let us assume $\Phi(\mathbf{X})$ to be zero mean. If $\Psi(\mathbf{X})$ stands for non-zero mean version of mapped in-sample data we obtain $\Phi(\mathbf{X})$ as:

$$\Phi(\mathbf{X}) = \Psi(\mathbf{X})(\mathbf{I}_N - \mathbf{E}_N) \quad (2)$$

with $\mathbf{I}_N$ being the $N \times N$ identity matrix, and $\mathbf{E}_N \triangleq \frac{1}{N}\mathbf{1}_N \mathbf{1}_N^T$ is an $N \times N$ matrix with all the elements equal to $1/N$. Let $\mathbf{K} \triangleq \Phi(\mathbf{X})^T \Phi(\mathbf{X}) \in \mathbb{R}^{N \times N}$ be a kernel (Gramm) matrix of the in-sample data. The rank of $\mathbf{K}$ is $R$. Because $\Phi(\mathbf{X})$ is centered it applies $R \le N-1$. Let $\mathcal{K} \triangleq \Psi(\mathbf{X})^T \Psi(\mathbf{X})$ be non-centered kernel matrix. We obtain centered kernel matrix as:

$$\mathbf{K} = (\mathbf{I}_N - \mathbf{E}_N)\mathcal{K}(\mathbf{I}_N - \mathbf{E}_N) \quad (3)$$

Let kernel vector associated with arbitrary $\mathbf{x} \in \mathbb{R}^D$ be defined as:

$$k(\mathbf{x}) \triangleq \Phi(\mathbf{X})^T \phi(\mathbf{x}) \in \mathbb{R}^N \quad (4)$$

Hence, $k(\mathbf{x})$ is EKM (1) with regard to in-sample dataset $\mathbf{X}$. Let $\kappa(\mathbf{x})$ be non-centered kernel vector. We obtain the centered one as:

$$k(\mathbf{x}) = (\mathbf{I}_N - \mathbf{E}_N)\left[\kappa(\mathbf{x}) - \frac{1}{N}\mathcal{K}\mathbf{1}_N\right]. \quad (5)$$

Let $\mathcal{P}$ be $R$-dimensional subspace of the feature space formed by $\Phi(\mathbf{X})$. Let $\phi_\mathbf{w}(\mathbf{x})$ be projection of $\phi(\mathbf{x})$ onto 1-D vector space formed by $\mathbf{w} \in \mathbb{R}^F$. We restrict $\mathbf{w}$ to be in $\mathcal{P}$, i.e. $\mathbf{w} = \Phi(\mathbf{X})\boldsymbol{\alpha}$ for some $\boldsymbol{\alpha} \in \mathbb{R}^N$. Let $\mathbf{K} = \mathbf{U}\boldsymbol{\Lambda}\mathbf{U}^T$ be eigenvalue decomposition of $\mathbf{K}$ composed of non-zero eigenvalues, i.e. $\mathbf{U} \in \mathbb{R}^{N \times R}$ and $\boldsymbol{\Lambda} = \text{diag}(\lambda_1,...,\lambda_R) \in \mathbb{R}^{R \times R}$. Then the orthonormal basis of $\mathcal{P}$ is obtained as:

$$\boldsymbol{\Pi} \triangleq \Phi(\mathbf{X})\mathbf{U}\boldsymbol{\Lambda}^{-1/2} = [\boldsymbol{\pi}_1,...,\boldsymbol{\pi}_R] \in \mathbb{R}^{F \times R} . \quad (6)$$

We obtain coordinates of the mapped training data $\Phi(\mathbf{X})$ in Cartesian coordinate system spanned by $\boldsymbol{\Pi}$ as:

$$\mathbf{Y} = \boldsymbol{\Lambda}^{-1/2}\mathbf{U}^T\mathbf{K} = \boldsymbol{\Lambda}^{1/2}\mathbf{U}^T \in \mathbb{R}^{R \times N}. \quad (7)$$

We obtain a coordinate of projection of any $\mathbf{x} \in \mathbb{R}^D$ onto $\mathcal{P}$ in the same coordinate system as:

$$\mathbf{y} = \boldsymbol{\Pi}^T\phi(\mathbf{x}) = \boldsymbol{\Lambda}^{-1/2}\mathbf{U}^T k(\mathbf{x}). \quad (8)$$

By using lemma 1 from [18], we obtain coordinates of projection vector $\mathbf{w}$ in $\mathcal{P}$ as:

$$\mathbf{w} = \Phi(\mathbf{X})\boldsymbol{\alpha} = \boldsymbol{\Pi}\boldsymbol{\beta} \quad (9)$$

where $\boldsymbol{\beta} = \mathbf{Y}\boldsymbol{\alpha} \in \mathbb{R}^R$. Given the fact that we now have the coordinates of the in-sample data (7), and out-of-sample (test) data (8) in the empirical feature space, applying the kernel method to in-sample data $\mathbf{X}$ is equivalent to applying the linear method to their coordinates $\mathbf{Y}$. The same applies to out-of-sample data $\mathbf{x}$ and its coordinates $\mathbf{y}$. In that regard we point out that EKM (8) is the same as the KPCA transform given by eq.(34) in [17]. In other words, KPCA can be implemented in a new more interpretable way by applying PCA on coordinates (7)/(8) [18].

## III. KERNEL SPARSE SUBSPACE CLUSTERING

### A. Robust Kernel Sparse Subspace Clustering

We now present proof of equivalence related to applying the NPT concept to robust SSC algorithm with the $\ell_1$-norm based error term [5]. In [5] SSC was proposed for solving the optimization problem:

$$\min_{\mathbf{C}} \|\mathbf{C}\|_1 + \lambda_e \|\mathbf{E}\|_1 + \frac{\lambda_z}{2}\|\mathbf{Z}\|_F^2 \\ \text{s.t.} \mathbf{X} = \mathbf{XC} + \mathbf{E} + \mathbf{Z}, \; \text{diag}(\mathbf{C}) = 0. \quad (10)$$

By setting $\lambda_z = 0$ we obtain robust SSC objective function:

$$\min_{\mathbf{C}} \|\mathbf{C}\|_1 + \lambda_e \|\mathbf{X} - \mathbf{XC}\|_1 \\ \text{s.t.} \mathbf{X} = \mathbf{XC} + \mathbf{E}, \; \text{diag}(\mathbf{C}) = 0. \quad (11)$$

Let us now do substitution $\mathbf{X} \to \Phi(\mathbf{X})$ in (11):

$$\min_{\mathbf{C}} \|\mathbf{C}\|_1 + \lambda_e \|\Phi(\mathbf{X}) - \Phi(\mathbf{X})\mathbf{C}\|_1 \\ \text{s.t.} \Phi(\mathbf{X}) = \Phi(\mathbf{X})\mathbf{C} + \mathbf{E}, \; \text{diag}(\mathbf{C}) = 0. \quad (12)$$

Since error term cannot be expressed in terms of inner product $\Phi(\mathbf{X})^T\Phi(\mathbf{X})$, (12) cannot be kernelized by using the *kernel trick*. We now re-formulate optimization problem (12) in projeted 1-D vectors space:

$$\min_{\mathbf{C},\mathbf{w}} \|\mathbf{C}\|_1 + \lambda_e \|\mathbf{w}^T(\Phi(\mathbf{X})-\Phi(\mathbf{X})\mathbf{C})\|_1$$
$$\text{s.t.} \Phi(\mathbf{X}) = \Phi(\mathbf{X})\mathbf{C}+\mathbf{E},\ \text{diag}(\mathbf{C})=0\ \text{and}\ \|\mathbf{w}\|_2 = 1.$$
(13)

By using (9) and the fact that $\Pi$ is orthonormal it follows $\|\mathbf{w}\|_2 = 1 \Rightarrow \|\boldsymbol{\beta}\|_2 = 1$. Now, bearing in mind that $\mathbf{w}^T\Phi(\mathbf{X}) = \boldsymbol{\beta}\mathbf{Y}$ [17], (13) becomes:

$$\min_{\mathbf{C},\boldsymbol{\beta}} \|\mathbf{C}\|_1 + \lambda_e \|\boldsymbol{\beta}^T(\mathbf{Y}-\mathbf{Y}\mathbf{C})\|_1$$
$$\text{s.t.}\ \mathbf{Y} = \mathbf{Y}\mathbf{C}+\mathbf{E},\ \text{diag}(\mathbf{C})=0\ \text{and}\ \|\boldsymbol{\beta}\|_2 = 1.$$

By using the matrix norm inequality $\|\mathbf{Ab}\|_1 \leq \|\mathbf{A}\|_1 \|\mathbf{b}\|_1$, eq. (2.3.4) in [20], we obtain:

$$\min_{\mathbf{C},\boldsymbol{\beta}} \|\mathbf{C}\|_1 + \lambda_e \|(\mathbf{Y}-\mathbf{Y}\mathbf{C})\|_1 \|\boldsymbol{\beta}\|_1$$
$$\text{s.t.}\ \mathbf{Y} = \mathbf{Y}\mathbf{C}+\mathbf{E},\ \text{diag}(\mathbf{C})=0\ \text{and}\ \|\boldsymbol{\beta}\|_2 = 1.$$

By using the inequality between vector norms $\|\boldsymbol{\beta}\|_2 \leq \|\boldsymbol{\beta}\|_1 \leq \sqrt{R}\|\boldsymbol{\beta}\|_2$, eq. (2.2.5) in [20], we obtain:

$$\min_{\mathbf{C},\boldsymbol{\beta}} \|\mathbf{C}\|_1 + \sqrt{R}\lambda_e \|(\mathbf{Y}-\mathbf{Y}\mathbf{C})\|_1 \|\boldsymbol{\beta}\|_2$$
$$\text{s.t.}\ \mathbf{Y} = \mathbf{Y}\mathbf{C}+\mathbf{E},\ \text{diag}(\mathbf{C})=0\ \text{and}\ \|\boldsymbol{\beta}\|_2 = 1.$$
(14)

Since minimum of (14) is constrained by $\|\boldsymbol{\beta}\|_2 = 1$, from the optimization point of view we obtain equivalent formulation of (14) as:

$$\min_{\mathbf{C}} \|\mathbf{C}\|_1 + \sqrt{R}\lambda_e \|(\mathbf{Y}-\mathbf{Y}\mathbf{C})\|_1$$
$$\text{s.t.}\ \mathbf{Y} = \mathbf{Y}\mathbf{C}+\mathbf{E},\ \text{diag}(\mathbf{C})=0.$$
(15)

Since optimization problem (15) is basically the same as (11), robust SSC algorithm can be directly applied to $\mathbf{Y}$ to obtain robust KSSC algorithm. Difference between (11) and (15) is that in (15) regularization coefficient is proportional with square root of the rank of the kernel matrix $\mathbf{K}$. Obviously, $R$ together with the kernel parameters forms additional hyperparameters to be selected by cross-validation.

Robust KSSC algorithm yields estimate of the self-representation matrix $\mathbf{C}$ from which we calculate data affinity matrix:

$$\mathbf{A} = \frac{|\mathbf{C}|+|\mathbf{C}|^T}{2}.$$
(16)

From (16) normalized graph Laplacian matrix is computed [21]:

$$\mathbf{L} = \mathbf{I} - (\mathbf{D})^{-1/2}\mathbf{A}(\mathbf{D})^{-1/2}$$
(17)

with the elements of diagonal degree matrix: $\mathbf{D}_{ii} = \sum_{j=1}^{N} \mathbf{A}_{ij}$. Spectral clustering algorithm [22], is applied to $\mathbf{L}$ to assign the cluster labels to data points: $\mathbf{F} \in \mathbb{N}_0^{N\times c}$, where $c$ stands for number of clusters. We summarize robust KSSC (RKSSC) algorithm in Algorithm 1.

---
**Algorithm 1**: Robust KSSC (RKSSC)
---

**Input**: Data $\mathbf{X} \in \mathbb{R}^{D\times N}$, number of clusters $c$, $\lambda_e = f(R)$, $R$, parameters of selected kernel function $\kappa(\cdot,\cdot)$.

**Output**: Assigned cluster indicator matrix $\mathbf{F} \in \mathbb{N}_0^{N\times k}$.

**Step 1**: Compute uncentered kernel matrix $\{[\mathcal{K}]_{ij} = \kappa(\mathbf{x}_i, \mathbf{x}_j)\}_{i,j=1}^{N}$.

**Step 2**: Compute centered kernel matrix through: $\mathbf{K} = (\mathbf{I}_N - \mathbf{E}_N)\mathcal{K}(\mathbf{I}_N - \mathbf{E}_N)$.

**Step 3**: Compute eigenvalue decomposition of $\mathbf{K}$ such that $\mathbf{K} \leftarrow \mathbf{U\Lambda U}^T$, with $\Lambda = \text{diag}(\lambda_1,...,\lambda_R)$ leading eigenvalues and $\mathbf{U}$ contains corresponding $R$ eigenvectors.

**Step 4**: Compute coordinates: $\mathbf{Y} = \Lambda^{1/2}\mathbf{U}^T$.

**Step 5**: Apply robust SSC algorithm [5] to $\mathbf{Y}$ to estimate self-representation matrix $\mathbf{C}$.

**Step 6**: Calculate data affinity matrix $\mathbf{A}$ (16), normalized Laplacian matrix $\mathbf{L}$ (17), and apply spectral clustering on $\mathbf{L}$ to assign cluster labels to data points: $\mathbf{F} \in \mathbb{N}_0^{N\times c}$.

*B. Non-robust Kernel Sparse Subspace Clustering*

Although, kernel SSC algorithm for the case $\lambda_e = 0$ is derived in [9], we formulate herein objective function using the NPT-approach. Optimization problem in the input data space becomes:

$$\min_{\mathbf{C}} \|\mathbf{C}\|_1 + \frac{\lambda_z}{2}\|\mathbf{X}-\mathbf{XC}\|_F^2$$
$$\text{s.t.}\ \mathbf{X} = \mathbf{XC}+\mathbf{Z},\ \text{diag}(\mathbf{C})=0.$$
(18)

Formulation of (18) in projected 1-D space yields:

$$\min_{\mathbf{C},\mathbf{w}} \|\mathbf{C}\|_1 + \frac{\lambda_z}{2}\|\mathbf{w}^T(\Phi(\mathbf{X})-\Phi(\mathbf{X})\mathbf{C})\|_F^2$$
$$\text{s.t.}\ \Phi(\mathbf{X}) = \Phi(\mathbf{X})\mathbf{C}+\mathbf{Z},\ \text{diag}(\mathbf{C})=0\ \text{and}\ \|\mathbf{w}\|_2 = 1.$$
(19)

Bearing in mind that $\mathbf{w}^T\Phi(\mathbf{X}) = \boldsymbol{\beta}\mathbf{Y}$ and inequality $\|\mathbf{Ab}\|_F^2 \leq \|\mathbf{A}\|_F^2 \|\mathbf{b}\|_F^2$ for some arbitrary $\mathbf{A}$ and $\mathbf{b}$, optimization of (19) is substituted by:

$$\min_{\mathbf{C},\boldsymbol{\beta}} \|\mathbf{C}\|_1 + \frac{\lambda_z}{2}\|(\mathbf{Y}-\mathbf{Y}\mathbf{C})\|_F^2 \|\boldsymbol{\beta}\|_2^2$$
$$\text{s.t.}\ \mathbf{Y} = \mathbf{Y}\mathbf{C}+\mathbf{Z},\ \text{diag}(\mathbf{C})=0\ \text{and}\ \|\boldsymbol{\beta}\|_2 = 1.$$
(20)

Since minimum of (20) is constrained by $\|\boldsymbol{\beta}\|_2 = 1$, from the optimization point of view we obtain equivalent formulation of (20) as:

$$\min_{\mathbf{C}} \|\mathbf{C}\|_1 + \frac{\lambda_z}{2}\|(\mathbf{Y}-\mathbf{YC})\|_F^2 \qquad (21)$$
$$\text{s.t. } \mathbf{Y} = \mathbf{YC}+\mathbf{Z}, \ \text{diag}(\mathbf{C}) = 0.$$

Since (21) is the same as (18), SSC algorithm can be directly applied to $\mathbf{Y}$ to obtain KSSC algorithm.

*C. (Robust) KSSC for out-of-sample problem*

The centered kernel vector $\mathbf{y}=k(\mathbf{x})$ of some arbitrary out-of-sample (test) data point $\mathbf{x}$, is given by (5). For out-of-sample data point, we do not want to re-run the (R)KSSC algorithm on the augmented in-sample data set again. Instead of that, and based on cluster labels obtained from Algorithm 1, we obtain partitions of in-sample data $\mathbf{Y}$:

$$\left\{ \mathbf{Y}_m \leftarrow \mathbf{Y}_m - \underbrace{[\overline{\mathbf{y}}_m ... \overline{\mathbf{y}}_m]}_{N_m \text{ times}} \right\}_{m=1}^c \quad \overline{\mathbf{y}}_m = \frac{1}{N_m}\sum_{n=1}^{N_m}\mathbf{Y}_m(n)$$
$$\bigcup_{m=1}^c \mathbf{Y}_m = \mathbf{Y}, \text{ and } \sum_{m=1}^c N_m = N. \qquad (22)$$

From $\{\mathbf{Y}_m = \mathbf{U}_m\Sigma_m\mathbf{V}_m^{\mathrm{T}}\}_{m=1}^c$ we estimate orthonormal bases from the first $d$ left singular vectors of partitions, i.e. $\{\mathbf{U}_m \in \mathbb{R}^{D\times d}\}_{m=1}^c$ [23]. We assign label $\{m\}_{m=1}^c$ to the test point $\mathbf{y}=k(\mathbf{x})$, according to the minimum of a point-to-a-subspace distance criterion [23]:

$$[\pi(\mathbf{y})]_m = \begin{cases} 1, & \text{if } m = \arg\min_{l\in\{1,...,c\}}\left\|\tilde{\mathbf{y}}^l - \mathbf{U}_l(\mathbf{U}_l)^{\mathrm{T}}\tilde{\mathbf{y}}^l\right\|_2 \\ 0, & \text{otherwise.} \end{cases}$$
(23)

where $\tilde{\mathbf{y}}^l = \mathbf{y} - \overline{\mathbf{y}}_l$.

IV. EXPERIMENTAL RESULTS

We compared proposed robust RKSSC and non-robust KSSC algorithms with corresponding linear SSC algorithms [5] as baselines. For this purpose we used the well-known Extended Yaleb (EYaleb) [24] and MNIST [25] datasets. EYaleb dataset contains 38 groups of face images, whereat each group contains 64 images of the same subject. In our experiment, we cluster all 38 groups, i.e. $c$=38. MNIST dataset contains 10 groups of handwritten digits with 1000 images in each group. In our experiment, we cluster all 10 groups, i.e. $c$=10. Regarding kernel function, we used polynomial kernel: $\kappa_{poly}(\mathbf{x}_i,\mathbf{x}_j) = (\langle\mathbf{x}_i,\mathbf{x}_j\rangle + b)^d$, and Gaussian kernel: $\kappa_{Gauss}(\mathbf{x}_i,\mathbf{x}_j) = \exp^{-\|\mathbf{x}_i-\mathbf{x}_j\|_2^2/(2\sigma^2)}$. Hereby, together with the kernel rank $R$, $(b, d)$ or $\sigma^2$ are hyperparameters introduced by kernelization. To tune hyperparameters we randomly generated subsets containing 46 and 200 data samples per group from EYaleb and MNIST datasets in respective order. We used accuracy (ACC), normalized mutual information (NMI), and $F_1$ score as performance measures. We validated performance metrics on 100 randomly generated in-sample and out-of-sample subsets. In-sample subsets contained the same number of samples as for hyperparameters tuning. Out-of-sample subsets contained 19 and 200 data samples from EYaleb and MNIST datasets in respective order. In terms of preprocessing, we normalized all data to the unit-column $\ell_2$ norm prior to further processing. For EYaleb dataset, hyperparameters of RSSC and RKSSC with Gaussian and polynomial kernel are in respective order: $\lambda_e = 15$; $R$=1490, $\sigma^2$=800, $\lambda_e = 0.1$; $R$=600, $b$=1, $d$=1, $\lambda_e = 0.15$. For MNIST dataset, hyperparameters in the same order are: $\lambda_e = 6$; $R$=380, $\sigma^2$=0.9, $\lambda_e = 0.1789$; $R$=600, $b$=0, $d$=11, $\lambda_e = 2.5044$. We conducted statistical significance analysis by Wilcoxon sum rank test. Obtained results are shown in Table I for EYaleb dataset, and in Table II for MNIST dataset. Performance on in-sample data is given in the first row, and out-of-sample data in the second row. RKSSC yielded statistically significantly better results than KSSC algorithm for both datasets. In case of EYaleB dataset, results obtained by RKSSC algorithm are also better from those obtained by the KSSC algorithm. To further emphasize quality of the results obtained by the RKSSC algorithm, we report results achieved by autoencoder (AE)-based networks deep networks [26], [27] in Table I, and for deep embedding for clustering (DEC) network [28] and deep subspace clustering network constrained by $\ell_2$ norm (DSC L2) [29] in Table II.

TABLE I: Clustering performance on EYaleb dataset.

| Algorithm | ACC [%] | NMI [%] | $F_1$[%] |
|---|---|---|---|
| RSSC | 75.65±1.84 | 80.48±1.65 | 40.77±4.39 |
|  | 81.36±2.17 | 85.79±1.67 | 58.73±3.81 |
| RKSSC Gauss | 80.63±1.78 | 84.78±0.09 | 68.01±2.63 |
|  | 81.72±1.84 | 86.62±1.09 | 70.73±2.49 |
| p values vs. RSSC | 4.41×10$^{-30}$ | 1.05×10$^{-33}$ | 2.56×10$^{-34}$ |
|  | 0.1801 | 1.33×10$^{-4}$ | 7.99×10$^{-34}$ |
| RKSSC poly | 81.74±1.73 | 85.59±0.91 | 70.51±2.59 |
|  | 82.03±1.83 | 86.97±1.12 | 71.95±2.59 |
| p values vs. RSSC | 2.28×10$^{-32}$ | 3.56×10$^{-34}$ | 2.56×10$^{-34}$ |
|  | 0.0435 | 5.54×10$^{-8}$ | 4.02×10$^{-34}$ |
| SSC | 62.34±1.55 | 68.12±1.01 | 27.96±1.83 |
|  | 71.30±2.19 | 77.47±1.55 | 46.83±3.38 |
| KSSC Gauss | 71.17±1.51 | 75.78±1.06 | 42.42±2.14 |
|  | 77.63±1.84 | 82.76±1.20 | 59.00±2.97 |
| p values vs. SSC | 2.54×10$^{-34}$ | 2.56×10$^{-34}$ | 2.56×10$^{-34}$ |
|  | 4.82×10$^{-32}$ | 1.41×10$^{-33}$ | 1.11×10$^{-33}$ |
| KSSC poly | 70.72±1.49 | 75.14±1.04 | 40.84±2.12 |
|  | 77.25±1.83 | 82.14±1.26 | 57.11±3.47 |
| p values vs. SSC | 2.61×10$^{-34}$ | 2.56×10$^{-34}$ | 2.56×10$^{-34}$ |
|  | 1.72×10$^{-31}$ | 5.32×10$^{-33}$ | 1.07×10$^{-31}$ |
| AE [26] | 88.75 | 87.53 | - |
| AE [27] | 84.73 | 86.75 | - |

TABLE II: Clustering performance on MNIST dataset.

| Algorithm | ACC [%] | NMI [%] | $F_1$[%] |
|---|---|---|---|
| RSSC | 60.25±4.89 | 62.10±2.95 | 51.31±3.86 |
|  | 60.30±4.35 | 60.40±2.36 | 52.52±3.43 |
| RKSSC Gauss | 64.57±2.72 | 62.94±2.03 | 64.57±2.72 |
|  | 65.18±2.83 | 64.02±2.09 | 65.18±2.83 |
| p values vs. RSSC | 2.05×10$^{-11}$ | 3.09×10$^{-2}$ | 9.29×10$^{-18}$ |
|  | 1.13×10$^{-13}$ | 6.56×10$^{-20}$ | 4.50×10$^{-17}$ |
| RKSSC poly | 62.49±2.48<br>59.27±2.84 | 62.11±1.72<br>58.64±1.89 | 54.44±2.19<br>47.83±2.70 |
| p values vs. RSSC | $9.97 \times 10^{-5}$<br>0.1136 | 0.9971<br>$1.68 \times 10^{-7}$ | $2.46 \times 10^{-10}$<br>$5.57 \times 10^{-18}$ |
| SSC | 59.72±3.72<br>60.02±3.30 | 59.22±2.77<br>58.53±2.07 | 49.07±3.34<br>51.92±2.84 |
| KSSC Gauss | 63.68±3.53<br>63.83±3.55 | 63.53±2.28<br>63.45±2.33 | 55.76±2.27<br>55.89±2.91 |
| p values vs. SSC | $1.68 \times 10^{-11}$<br>$1.41 \times 10^{-9}$ | $1.18 \times 10^{-6}$<br>$9.07 \times 10^{-15}$ | $3.00 \times 10^{-15}$<br>$4.83 \times 10^{-10}$ |
| KSSC poly | 61.36±6.11<br>61.65±5.87 | 63.14±3.08<br>62.36±3.36 | 61.36±6.11<br>61.65±5.87 |
| p values vs. SSC | $2.29 \times 10^{-4}$<br>0.0021 | $5.57 \times 10^{-5}$<br>$4.67 \times 10^{-7}$ | $1.65 \times 10^{-10}$<br>$5.02 \times 10^{-5}$ |
| DEC [28] | 61.20 | 57.53 | - |
| DSC l2 [29] | 75.00 | 73.19 | - |

## V. Conclusion

In this paper, we introduced robust kernel sparse subspace clustering (RKSSC) algorithm for nonlinear SC of data contaminated by gross sparse corruptions or outliers. In such a case, the error term in related linear optimization problem is modeled by the $\ell_1$-norm, and *kernel trick* based kernelization approach cannot be applied. By centering data in empirical feature space (EFS) induced by in kernel-based mapping, usage of nonlinear projection trick enables to calculate coordinates of mapped data with respect to orthonormal basis of a subspace of EFS. In such a scenario, we have shown equivalence between objectives functions of RSSC in the input data space and EFS. Thus, RKSSC is obtained by applying existing linear RSSC to the coordinates of data mapped in EFS, and that is equivalent to applying RKSSC method (that cannot be derived) to data in original input space. In our future work, we plan to derive nonlinear kernel-based versions of several others robust SC algorithms.

## Acknowledgment

This work was supported by the Croatian science foundation grant IP-2022-10-6403.